\documentclass[runningheads]{llncs}
\usepackage{graphicx}
\usepackage{amsmath,amssymb} 
\usepackage{color}
\usepackage[width=122mm,left=12mm,paperwidth=146mm,height=193mm,top=12mm,paperheight=217mm]{geometry}
\usepackage{cite}
\usepackage[]{hyperref}
\usepackage[utf8x]{inputenc}
\usepackage{footnote}
\makesavenoteenv{tabular}
\makesavenoteenv{table}

\newcommand{\norm}[1]{\left\lVert#1\right\rVert}

\begin{document}
\pagestyle{headings}
\mainmatter
\def\ECCV18SubNumber{757}  

\title{MultiPoseNet: Fast Multi-Person Pose Estimation using Pose Residual Network}

\titlerunning{ }
\authorrunning{ }

\author{Muhammed Kocabas$^1$, M. Salih Karagoz$^1$, Emre Akbas$^1$}
\institute{$^1$Department of Computer Engineering, Middle East Technical University \\ \texttt{\{muhammed.kocabas,e234299,eakbas\}@metu.edu.tr}}

\maketitle

\begin{abstract}
In this paper, we present \textit{MultiPoseNet}, a novel bottom-up multi-person pose estimation architecture that combines a multi-task model with a novel assignment method.  MultiPoseNet can jointly handle person detection, keypoint detection, person segmentation and pose estimation problems. The novel assignment method is implemented by the \textit{Pose Residual Network (PRN)} which receives keypoint and person detections, and produces accurate poses by assigning keypoints to person instances. On the COCO keypoints dataset, our pose estimation method outperforms all previous bottom-up methods both in accuracy (+4-point mAP over previous best result) and speed; it also performs on par with the best top-down methods while being at least 4x faster.  Our method is the fastest real time system with $\sim$23 frames/sec. Source code is available at: \url{https://github.com/mkocabas/pose-residual-network}


\end{abstract}

\section{Introduction}
This work is aimed at estimating  the two-dimensional (2D) poses of multiple people in a given image. Any solution to this problem has to tackle a few sub-problems: detecting body joints (or keypoints\footnote{We use ``body joint’’ and ``keypoint’’ interchangeably throughout the paper.}, as they are called in the influential COCO \cite{Lin2014} dataset) such as wrists, ankles, etc., grouping these joints into person instances, or detecting people and assigning joints to person instances. Depending on which sub-problem is tackled first, there have been two major approaches in multi-person 2D estimation: \textit{bottom-up} and \textit{top-down}. Bottom-up methods \cite{Cao2016, Pishchulin2015a, Insafutdinov2016, Bulata, Iqbal2016a, Ning2017, Newell2016b} first  detect body joints without having any knowledge as to the number of people or their locations. Next, detected joints are  grouped to form individual poses for person instances. On the other hand, top-down methods \cite{Chen2017a, Papandreou2017, He2017a, Fang2017} start by detecting people first and then for each person detection, a single-person pose estimation method (e.g. \cite{Wei2016, Newella, Chou2017, Huang}) is executed. Single-person pose estimation, i.e. detecting body joints conditioned on the information that there is a single person in the given input (the top-down approach), is typically a more costly process than grouping the detected joints  (the bottom-up approach). Consequently, the top-down methods tend to be slower than the bottom-up methods, since they need to repeat the single-person pose estimation for each person detection; however, they usually yield better accuracy than bottom-up methods.

\begin{figure}
\centering
\includegraphics[width=\textwidth]{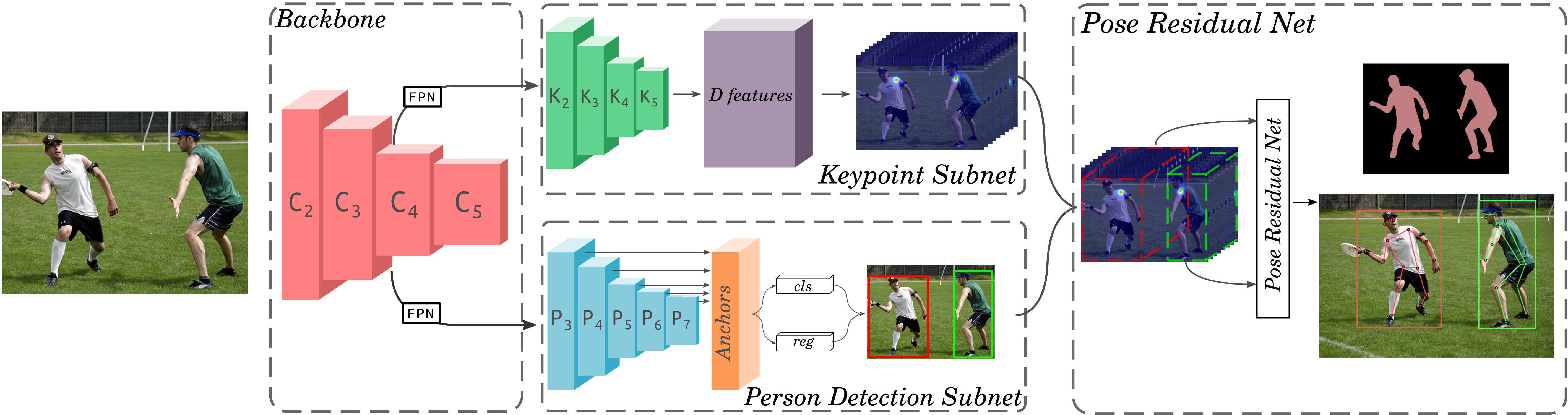}
\caption{MultiPoseNet is a multi-task learning architecture capable of performing human keypoint estimation, detection and semantic segmentation tasks altogether efficiently.}
\label{fig:arch}
\end{figure}

In this paper, we present a new bottom-up method for multi-person 2D pose estimation. Our method is based on a multi-task learning model which can jointly handle the person detection, keypoint detection, person segmentation and pose estimation problems. To emphasize its multi-person and multi-task aspects of our model, we named it as ``MultiPoseNet.’’ Our model (Fig. \ref{fig:arch}) consists of a shared backbone for feature extraction, detection subnets for keypoint and person detection/segmentation, and a final network which carries out the pose estimation, i.e. assigning detected keypoints to person instances.

\begin{figure}
\centering
\includegraphics[width=.8\textwidth]{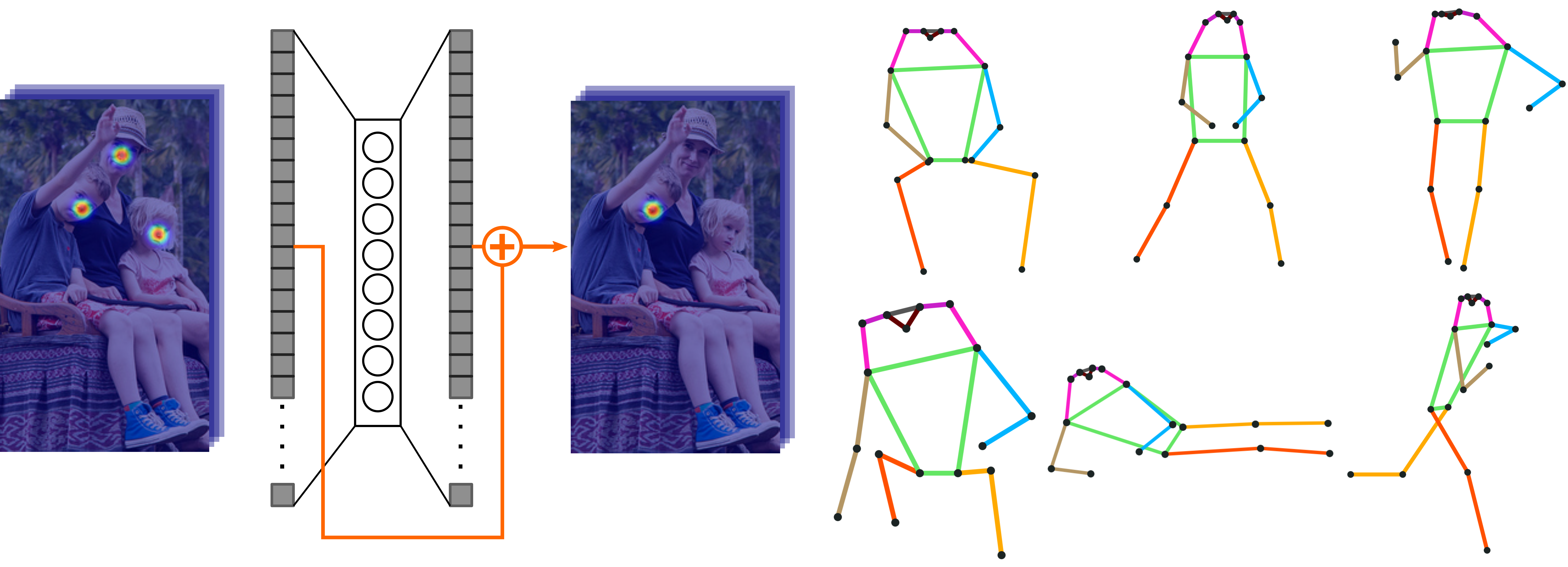}
\caption{\textbf{Left:} Pose Residual Network (PRN). The PRN is able to disambiguate which keypoint should be assigned to the current person box. \textbf{Right:} Six sample poses obtained via clustering the structures learned by PRN.}
\label{fig:psnatom}
\end{figure}

Our major contribution lies in the pose estimation step where the network implements a novel assignment method. This network receives keypoint and person detections, and produces a pose for each detected person by assigning keypoints to person boxes using a learned function. In order to put our contribution into context, here we briefly describe the relevant aspects of the state-of-the-art (SOTA) bottom-up methods \cite{Cao2016,Newell2016b}.  These methods attempt to group detected keypoints by exploiting lower order relations either between the group and keypoints, or among the keypoints themselves. Specifically, Cao et al. \cite{Cao2016} model pairwise relations (called part affinity fields) between two nearby joints and the grouping is achieved by propagating these pairwise affinities. In the other SOTA method, Newell et al. \cite{Newell2016b} predict a real number called a \textit{tag} per detected keypoint, in order to identify the group the detection belongs to. Hence, this model makes use of the unary relations between a certain keypoint and the group it belongs to. Our method generalizes these two approaches in the sense that we achieve the grouping in a single shot by considering all joints together at the same time. We name this part of our model which achieves the grouping as the \textit{Pose Residual Network} (PRN) (Fig. \ref{fig:psnatom}). PRN takes a region-of-interest (RoI) pooled keypoint detections and then feeds them into a residual multilayer perceptron (MLP). PRN considers all joints simultaneously and learns configurations of joints. We illustrate this capability of PRN by plotting a sample set of learned configurations. (Fig. \ref{fig:psnatom} right).

Our experiments (on the COCO dataset, using no external data) show that our method outperforms all previous bottom-up methods: we achieve a 4-point mAP increase over the previous best result. Our method performs on par with the best performing top-down methods while being an order of magnitude faster than them. To the best of our knowledge, there are only two top-down methods that we could not outperform.  Given the fact that bottom-up methods have always performed less accurately than the top-down methods, our results are remarkable.

In terms of running time, our method appears to be the fastest of all multi-person 2D pose estimation methods. Depending on the number of people in the input image, our method runs at between $27$ frames/sec (FPS) (for one person detection) and $15$ FPS (for $20$ person detections). For a typical COCO image, which contains $\sim$3 people on average, we achieve $\sim$23 FPS (Fig. \ref{fig:runtime}).

Our contributions in this work are four fold. (1) We propose the \textit{Pose Residual Network} (PRN), a simple yet very effective method for the problem of assigning/grouping body joints. (2) We outperform all previous bottom-up methods and achieve comparable performance with top-down methods. (3) Our method works faster than all previous methods, in real-time at $\sim$23 frames/sec. (4) Our network architecture is extendible; we show that using the same backbone, one can solve other related problems, too, e.g. person segmentation.
 
\section{Related Work}
\subsection{Single Person Pose Estimation}
Single person pose estimation is to predict individual body parts given  a cropped person image (or, equivalently, given its exact location and scale within an image). Early methods (prior to deep learning) used hand-crafted HOG features \cite{Dalal2005} to detect body parts and probabilistic graphical models to represent the pose structure (tree-based  \cite{Pishchulin2013,Yang2011,Johnson,Andriluka2009}; non-tree based \cite{Dantone2013,Gkioxari2014}).


Deep neural networks based models \cite{Toshev,Tompson,Carreira2015,Wei2016,Newella,Chu2017,Yang2011,Lifshitz2016,Huang,Belagiannis} have quickly dominated the pose estimation problem after the initial work by Toshev et al. \cite{Toshev} who used  the AlexNet architecture to directly regress spatial joint coordinates. Tompson et al. \cite{Tompson} learned pose structure by combining deep features along with graphical models. Carreira et al. \cite{Carreira2015} proposed the Iterative Error Feedback method to train Convolutional Neural Networks (CNNs) where the input is repeatedly fed to the network along with current predictions in order to refine the predictions. Wei et al.\cite{Wei2016} were inspired by the pose machines \cite{Ramakrishna2014} and used CNNs as feature extractors in pose machines.  \textit{Hourglass blocks}, (HG) developed by Newell et al. \cite{Newella}, are basically  convolution-deconvolution structures with  residual connections. Newell et al. stacked HG blocks to obtain an iterative refinement process and showed its effectiveness on single person pose estimation. Stacked Hourglass (SHG) based methods made a remarkable performance increase over previous results.  Chu et al. \cite{Chu2017} proposed  adding visual attention units to focus on keypoint regions of interest. Pyramid residual modules by Yang et al.\cite{Yang2011} improved the SHG architecture to handle  scale variations. Lifshitz et al. \cite{Lifshitz2016} used a probabilistic keypoint voting scheme from image locations to obtain agreement maps for each body part. Belagiannis et al. \cite{Belagiannis} introduced a simple recurrent neural network based prediction refinement architecture. Huang et al.\cite{Huang} developed a coarse-to-fine model with Inception-v2 \cite{Szegedy15} network as the backbone. The authors calculated the loss in each level of the network to learn coarser to finer representations of parts.

\subsection{Multi Person Pose Estimation}

\subsubsection{Bottom-up}
Multi person pose estimation solutions branched out as bottom-up and top-down methods. Bottom-up approaches detect body joints and assign them to people instances, therefore they are faster in test time and smaller in size compared to top-down approaches. However, they miss the opportunity to zoom into the details of each person instance. This creates an accuracy gap between top-down and bottom-up approaches.

In an earlier work by Ladicky et al.\cite{LadickyETHZurich}, they proposed an algorithm to jointly predict human part segmentations and part locations using HOG-based features and probabilistic approach. Gkioxari et al. \cite{Gkioxari2013} proposed k-poselets to jointly detect people and keypoints.

Most of the  recent approaches use Convolutional Neural Networks (CNNs) to detect body parts and relationships between them in an end-to-end manner \cite{Cao2016, Newell2016b, Varadarajan2017, Pishchulin2015a, Pishchulin2013, Insafutdinov2016}, then use assignment algorithms \cite{Cao2016,Pishchulin2015a,Insafutdinov2016,Varadarajan2017} to form individual skeletons.

Pischulin et al.\cite{Pishchulin2015a} used deep features for joint prediction of part locations and relations between them, then performed correlation clustering. Even though \cite{Pishchulin2015a} doesn't use person detections, it is very slow due to proposed clustering algorithm and processing time is in the order of hours. In a following work by Insafutdinov et al.\cite{Insafutdinov2016}, they benefit from deeper ResNet architectures as part detectors and improved the parsing efficiency of a previous approach with an incremental optimization strategy. Different from Pischulin and Insafutdinov, Iqbal et al. \cite{Iqbal2016} proposed to solve the densely connected graphical model locally, thus improved time efficiency significantly.

Cao et al.\cite{Cao2016} built a model that contain two entangled CPM\cite{Wei2016} branches to predict keypoint heatmaps and pairwise relationships (part affinity fields) between them. Keypoints are grouped together with fast Hungarian bipartite matching algorithm according to conformity of part affinity fields between them. This model runs in realtime. Newell et al.\cite{Newell2016b} extended their SHG idea by outputting associative vector embeddings which can be thought as tags representing each keypoint's group. They group keypoints with similar tags into individual people.

\subsubsection{Top-down}

Top-down methods first detect people (typically using a top performing, off-the-shelf object detector) and then run a single person pose estimation (SPPN) method per person to get the final pose predictions. Since a SPPN model is run for each person instance, top-down methods are extremely slow, however, each pose estimator can focus on an instance and perform fine localization. Papandreou et al.\cite{Papandreou2017} used ResNet with dilated convolutions \cite{He2016} which has been very successful in semantic segmentation \cite{Chen2016} and computing keypoint heatmap and offset outputs. In contrast to Gaussian heatmaps, the authors estimated a disk-shaped keypoint masks and 2-D offset vector fields to accurately localize keypoints. Joint part segmentation and keypoint detection given human detections approach were proposed by Xia et al.\cite{Xia} The authors used separate PoseFCN and PartFCN to obtain both part masks and locations and fused them with fully-connected CRFs. This provides more consistent predictions by eliminating irrelevant detections. Fang et al.\cite{Fang2017} proposed to use spatial transformer networks to handle inaccurate bounding boxes and used \textit{stacked hourglass} blocks \cite{Newella}. He et al.\cite{He2017a} combined instance segmentation and keypoint prediction in their \textit{Mask-RCNN} model. They append keypoint heads on top of \textit{RoI aligned} feature maps to get  a one-hot mask for each keypoint. Chen et al.\cite{Chen2017a} developed \textit{globalnet} on top of \textit{Feature Pyramid Networks} \cite{Lina} for multiscale inference and refined the predictions by using hyper-features \cite{Kong}.
\section{The Method and Models} 
The architecture of our proposel model, MultiPoseNet, can be found in Fig. \ref{fig:arch}. In the following, we describe each component in detail.

\subsection{The Shared Backbone}
\label{sec:method-backbone}
The backbone of MultiPoseNet serves as  a feature extractor for keypoint and person detection subnets. It is actually a ResNet \cite{He2016} with two Feature Pyramid Networks (FPN)\cite{Lina} (one for the keypoint subnet, the other for the person detection subnet) connected to it, FPN creates pyramidal feature maps with top-down connections from all levels of CNN’s feature hierarchy to make use of inherent multi-scale representations of a CNN feature extractor. By doing so, FPN compromises high resolution, weak representations with low resolution, strong representations. Powerful localization and classification properties of FPN proved to be very successful in detection, segmentation and keypoint tasks recently \cite{Chen2017a, He2017a, Lin2017, Lina}. In our model, we extracted features from the last residual blocks $C_2, C_3, C_4, C_5$ with strides of (4,8,16,32) pixels and compute corresponding FPN features per subnet.
\subsection{Keypoint Estimation Subnet}
\label{sec:kp}
\begin{figure}
\centering
\includegraphics[width=\textwidth]{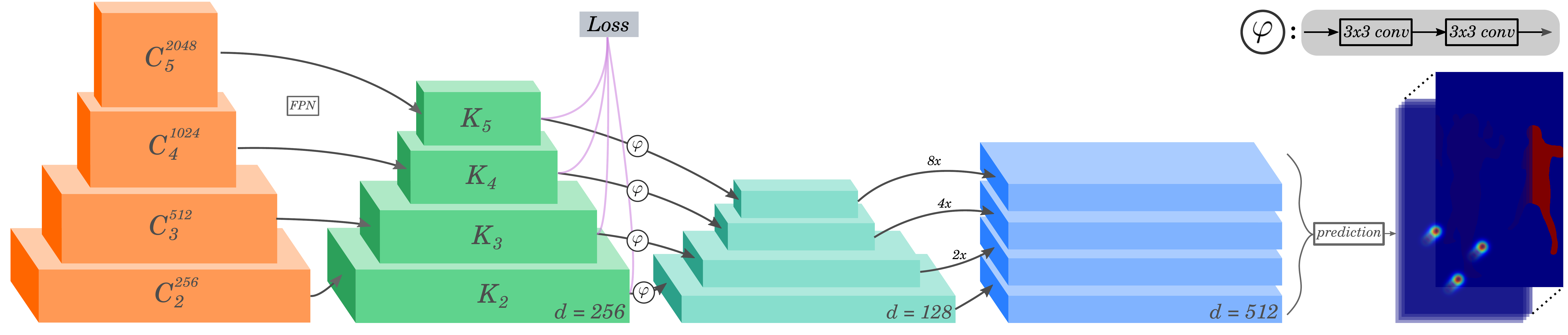}
\caption{The architecture of the keypoint subnet. It takes hierarchical CNN features as input and outputs keypoint and segmentation heatmaps.}
\label{fig:kp-arch}
\end{figure}

Keypoint estimation subnet (Fig. \ref{fig:kp-arch}) takes hierarchical CNN features (outputted by the corresponding FPN) and outputs keypoint and segmentation heatmaps. Heatmaps represent keypoint locations as Gaussian peaks. Each heatmap layer belongs to a specific keypoint class (nose, wrists, ankles etc.) and contains arbitrary number of peaks that pertain to person instances. Person segmentation mask at the last layer of heatmaps encodes the pixelwise spatial layout of people in the image.

A set of features specific to the keypoint detection task are computed similarly to \cite{Lina} with top-down and lateral connections from the bottom-up pathway. $K_2 - K_5$ features have the same spatial size corresponding to $C_2 - C_5$ blocks but the depth is reduced to 256. $K$ features are identical to $P$ features in the original FPN paper, but we denote them with  $K$ to distinguish from person detection subnet layers. The depth of P features is downsized to 128 with 2 subsequent $3 \times 3$ convolutions to obtain $D_2, D_3, D_4, D_5$ layers. Since $D$ features still have different strides, we upsampled $D_3, D_4, D_5$ accordingly to match 4-pixel stride as $D_2$ features and concatenated them into a single depth-512 feature map. Concatenated features are smoothed by a $3 \times 3$ convolution with ReLU. Final heatmap which has $(K+1)$ layers obtained via $1 \times 1$ convolutions without activation. The final output is multiplied with a binary mask of $\mathbf{W}$ which has $\mathbf{W}(\mathbf{p})=0$ in the area of the persons without annotation. $K$ is the number of human keypoints annotated in a dataset and $+1$ is person segmentation mask. In addition to the loss applied in the last layer, we append a loss at each level of $K$ features to benefit from intermediate supervision. Semantic person segmentation masks are predicted in the same way with keypoints.
\subsection{Person Detection Subnet}
Modern object detectors are classified as one-stage (SSD\cite{liu2016ssd}, YOLO\cite{redmon}, RetinaNet \cite{Lin2017}) or two-stage (Fast R-CNN\cite{girshick15fastrcnn}, Faster R-CNN\cite{ren2015faster}) detectors. One-stage detectors enable faster inference but have lower accuracy in comparison to two-stage detectors due to foreground-background class imbalance. The recently proposed RetinaNet \cite{Lin2017} model improved one-stage detectors'   performance with \textit{focal loss} which can handle the class imbalance problem during training. In order to design a faster and simpler person detection model which is compatible with FPN backbone, we have adopted RetinaNet. Same strategies to compute anchors, losses and pyramidal image features are followed. Classification and regression heads are modified to handle only person annotations.

\subsection{Pose Residual Network (PRN)}
\label{sec:method-prn}

\begin{figure}
\centering
\includegraphics[width=\textwidth]{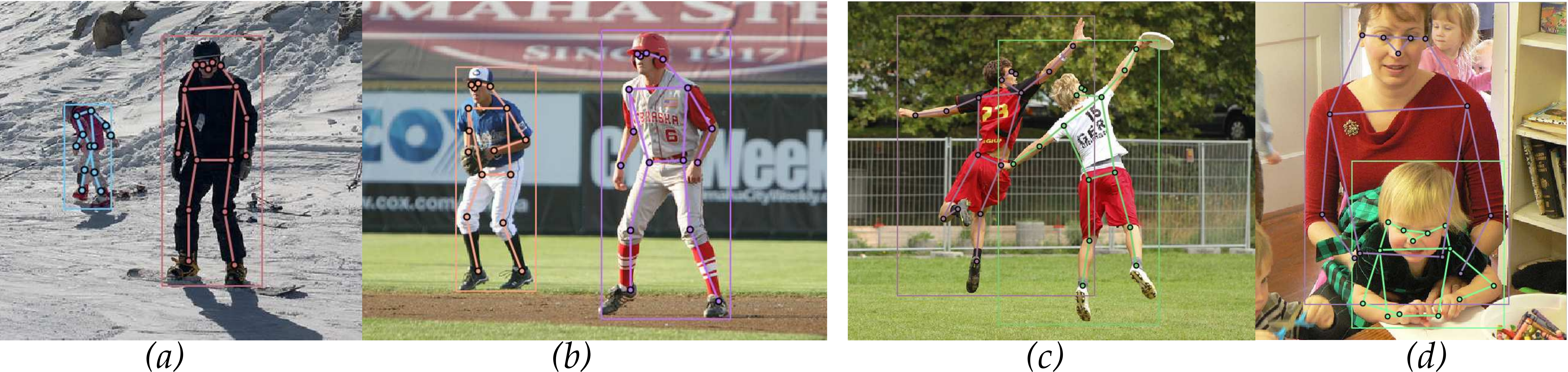}
\caption{Bounding box overlapping scenarios.}
\label{fig:overlap}
\end{figure}

Assigning keypoint detections to person instances (bounding boxes, in our case) is straightforward if there is only one person in the bounding box as in Fig. \ref{fig:overlap} a-b. However, it becomes non-trivial if there are overlapping people in a single box as in Fig. \ref{fig:overlap} c-d. In the case of an overlap, a bounding box can contain multiple keypoints not related to the person in question, so this creates ambiguity in constructing final pose predictions. We solve these ambiguities by learning pose structures from data.

The input to PRN is prepared as follows. For each person box that the person detection subnet detected, the region from the keypoint detection subnet's output, corresponding to the box, is cropped and resized to a fixed size, which ensures that PRN can handle person detections of arbitrary sizes and shapes. Specifically, let $\bf X$ denote the input to the PRN, where $\mathbf{X} = \{\mathbf{x}_1, \mathbf{x}_2, \dots, \mathbf{x}_k\}$ in which $\mathbf{x}_k \in \mathbb{R}^{W \times H}$, $k$ is the number of different keypoint types. The final goal of PRN is to output $\bf Y$ where $\mathbf{Y} = \{\mathbf{y}_1, \mathbf{y}_2, \dots, \mathbf{y}_k\}$, in which $\mathbf{y}_k \in \mathbb{R}^{W \times H}$ is of the same size as ${\bf x}_k$, containing the correct position for each keypoint indicated by a peak in that keypoint’s channel. PRN models the mapping from $\bf X$ to $\bf Y$ as

\begin{equation}
{\bf y}_k = \phi_k ( {\bf X}) + {\bf x}_k 
\label{eq:prn}
\end{equation}

\noindent where the functions $\phi_1(\cdot), \dots , \phi_K(\cdot)$ apply a \textit{residual correction} to the pose in $\bf X$, hence the name pose residual network. We implement Eq. \ref{eq:prn} using a residual multilayer perceptron Fig. \ref{fig:psnatom}. Activation of the output layer uses softmax to obtain a proper probability distribution and binary cross-entropy loss is used during training.

Before we came up with this residual model, we experimented with two naive baselines and a non-residual model. In the first baseline method, which we call \textit{Max}, for each keypoint channel $k$, we find the location with the highest value and place  a Gaussian in the corresponding location of the $k^{th}$ channel in $\bf Y$.  In the second baseline method, we compute $\bf Y$ as

\begin{equation}
{\bf y}_k = {\bf x}_k * {\bf P}_k
\end{equation}

\noindent where ${\bf P}_k$ is a prior map for the location of the $k^{th}$ joint, learned from ground-truth data and $*$ is element-wise multiplication. We named this method as Unary Conditional Relationship (UCR). Finally, in our non-residual model, we implemented

\begin{equation}
{\bf y}_k = \phi_k ( {\bf X}). 
\end{equation}

\noindent Performances of all these models can be found in Table   \ref{table:psn}.

In the context of the models described above, both SOTA bottom up methods learn lower order grouping models than the PRN. Cao et al. \cite{Cao2016} model pairwise channels in ${\bf X}$ while Newell et al. \cite{Newell2016b} model only unary channels in ${\bf X}$. Hence, our model can be considered as a generalization of these lower order grouping models.

We hypothesize that each node in PRN's hidden layer encodes a certain body configuration. To show this, we visualized some of the representative outputs of PRN in Fig. \ref{fig:psnatom}. These poses is obtained via reshaping PRN outputs and selecting the maximum activated keypoints to form skeletons. All obtained configurations are clustered using $k$-means with OKS (object keypoint similarity)\cite{Lin2014} and cluster means are visualized in Fig. \ref{fig:psnatom}. OKS (object keypoint similarity) is used as k-means distance metric to cluster the meaningful poses.
 


\subsection{Implementation Details} 
\subsubsection{Training}
Due to different convergence times and loss imbalance, we have trained keypoint and person detection tasks separately. To use the same backbone in both task, we first trained the model with only keypoint subnet Fig. \ref{fig:kp-arch}. Thereafter, we froze the backbone parameters and trained the person detection subnet. Since the two tasks are semantically similar, person detection results were not adversely affected by the frozen backbone.

We have utilized Tensorflow \cite{tensorflow2015-whitepaper} and Keras \cite{chollet2015keras} deep learning library to implement training and testing procedures. For person detection, we made use of open-source Keras RetinaNet\cite{hans_gaiser_2018_1188105} implementation.
\paragraph{Keypoint Estimation Subnet:} 
For keypoint training, we used $480$x$480$ image patches, that are centered around the crowd or the main person in the scene. Random rotations between $\pm 40$ degrees, random scaling between $0.8-1.2$ and vertical flipping with a probability of 0.3 was used during training. We have transferred the ImageNet \cite{imagenet} pretrained weights for each backbone before training. We optimize the model with Adam \cite{Kingma} starting from learning rate 1e-4 and decreased it by a factor of 0.1 in plateaux. We used the Gaussian peaks located at the keypoint locations as the ground truth to calculate $L_2$ loss, and we masked (ignored) people that are not annotated. We appended the segmentation masks to ground-truth as an extra layer and trained along with keypoint heatmaps. The cost function that we minimize is

\begin{equation}
L_{kp} = \mathbf{W} \cdot \norm{\mathbf{H}_t - \mathbf{H}_p}_{2}^{2},
\end{equation}

\noindent where $\mathbf{H}_t$ and $\mathbf{H}_p$ are the ground-truth and predicted heatmaps respectively, and $\mathbf{W}$ is the mask used to ignore non-annotated person instances. 
\paragraph{Person Detection Subnet:} 
We followed a similar person detection training strategy as \cite{Lin2017}. Images containing persons are used, they are resized such that shorter edge is 800 pixels. We froze backbone weights after keypoint training and not updated during person detection training. We optimized subnet with Adam \cite{Kingma} starting from learning rate 1e-5 and is decreased by a factor of 0.1 in plateaux. We used Focal loss with $(\gamma=2, \alpha=0.25)$ and smooth $L_1$ loss for classification and bbox regression, respectively. We obtained final proposals using NMS with a threshold of 0.3.
\paragraph{Pose Residual Network:}
During training, we cropped input and output pairs and resized heatmaps according to bounding-box proposals. All crops are resized to a fixed size of $36 \times 56$ (height/width = 1.56). We trained the PRN network separately and Adam optimizer \cite{Kingma} with a learning rate of 1e-4 is used during training. Since the model is shallow, convergence takes 1.5 hours approximately.

We trained the model with the person instances which has more than 2 keypoints. We utilized a sort  of curriculum learning \cite{bengio2009} by sorting annotations based on number of keypoints and bounding box areas. In each epoch, model is started to learn easy-to-predict instances, hard examples are given in later stages. 
\subsubsection{Inference}
The whole architecture (see in Fig. \ref{fig:arch}) behaves as  a monolithic, end-to-end model during test time. First, an image $(W \times H \times 3)$ is processed through backbone model to extract the features in multi-scales. Person and keypoint detection subnets compute outputs simultaneously out of extracted features. Keypoints are outputted as $W \times H \times (K+1)$ sized heatmaps. $K$ is the number of keypoint channels, and $+1$ is for the segmentation channel. Person detections are in the form of $N \times 5$, where $N$ is the number of people and 5 channel corresponds to 4 bounding box coordinates along with confidence scores. Keypoint heatmaps are cropped and resized to form RoIs according to person detections. Optimal RoI size is determined as $36 \times 56 \times (K+1)$ in our experiments. PRN takes each RoI as separate input, then outputs same size RoI with only one keypoint selected in each layer of heatmap. All selected keypoints are grouped as a person instance.

\section{Experiments}
\subsection{Datasets}
We trained our keypoint and person detection models on COCO keypoints dataset \cite{Lin2014} (without using any external/extra data) in our experiments. We used COCO for evaluating the keypoint and person detection, however, we used PASCAL VOC 2012\cite{Everingham15} for evaluating person segmentation due to the lack of semantic segmentation annotations in COCO. Backbone models (ResNet-50 and ResNet-101) were pretrained on ImageNet and we finetuned with COCO-keypoints.

COCO train2017 split contains 64K images including 260K person instances which 150K of them have keypoint annotations. Keypoints of persons with small area are not annotated in COCO. We did ablation experiments on COCO val2017 split which contains 2693 images with person instances. We made comparison to previous methods on the test-dev2017 split which has 20K test images. We evaluated test-dev2017 results on the online COCO evaluation server. We use the official COCO evaluation metric average precision (AP) and average recall (AR). OKS and IoU based scores were used for keypoint and person detection tasks, respectively.

We performed person segmentation evaluation in PASCAL VOC 2012 test split with PASCAL IoU metric. PASCAL VOC 2012 person segmentation test split contains 1456 images. We obtained ``Test results’’ using the online evaluation server.
\subsection{Multi Person Pose Estimation}
\begin{figure}
\centering
\includegraphics[width=\textwidth]{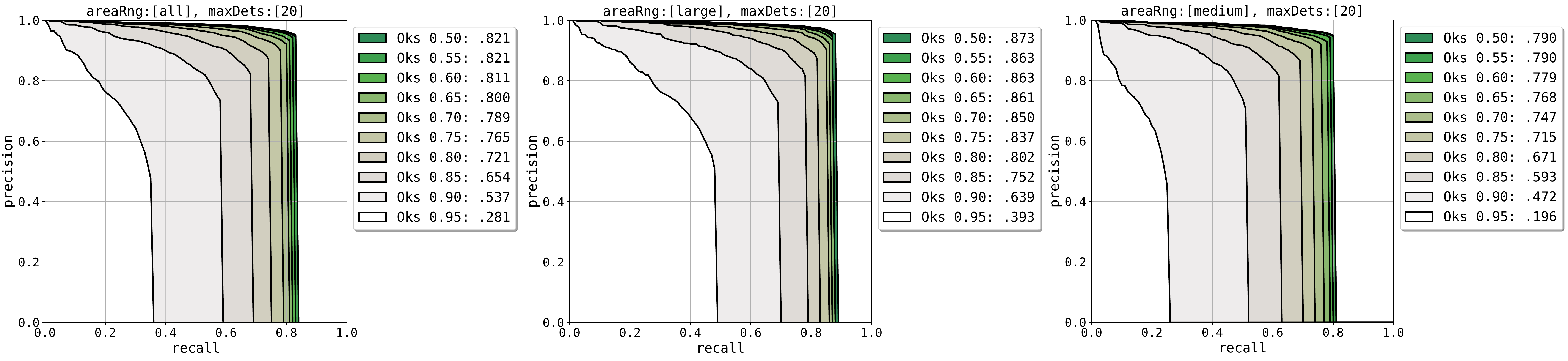}
\caption{Precision-recall curves on COCO validation set across scales \textit{all, large} and \textit{medium}. Analysis tool is provided by \cite{Ronchi2017}}
\label{fig:analyze}
\end{figure}

We present the recall-precision curves of our method for different scales \textit{all, large, medium} in Fig. \ref{fig:analyze}. The overall AP results of our method along with top-performing bottom-up (BU) and top-down (TD) methods are given in  Table \ref{table:test}.  MultiPoseNet outperforms all bottom-up methods and most of the top-down methods. We outperform the previously best bottom-up method\cite{Newell2016b} by a 4-point increase in mAP. In addition, the runtime speed (see the FPS column Table \ref{table:test} and Fig. \ref{fig:runtime}) of our system is far better than previous methods with 23 FPS on average\footnote{We obtained the FPS results by averaging the inference time using images containing 3 people (avg. number of person annotations per image in \href{http://image-net.org/challenges/talks/2016/ECCV2016_workshop_presentation_keypoint.pdf}{COCO dataset}) on a GTX1080Ti GPU. Except for CFN and Mask RCNN, we obtained the FPS numbers by running the models ourselves under equal conditions. CFN’s code is not available and Mask RCNN’s code was only made recently available and we did not have time to test it ourselves. We got CFN’s and Mask RCNN’s FPS from their respective papers.}. This proves the effectiveness of PRN for assignment and our multitask detection approach while providing reasonable speed-accuracy tradeoff. To get these results (Table \ref{table:test}) on test-dev, we have utilized test time augmentation and ensembling (as also done in all previous studies). Multi scale and multi crop testing was performed during test time data augmentation. Two different backbones and a single person pose refinement network similar to our keypoint detection model was used for ensembling. Results from different models are gathered and redundant detections was removed via OKS based NMS \cite{Papandreou2017}.

\begin{table}
\caption{Results on COCO \textbf{test-dev}, excluding systems trained with external data. Top-down methods are shown separately to make a clear comparison between bottom-up methods.}
\label{table:test}
\resizebox{\textwidth}{!}{\begin{tabular}{l||l||c|cccccccccc}
\hline
& & \textbf{FPS} & $\mathbf{AP}$ & $\mathbf{AP_{50}}$ & $\mathbf{AP_{75}}$ & $\mathbf{AP_{M}}$ & $\mathbf{AP_{L}}$ & $\mathbf{AR}$ & $\mathbf{AR_{50}}$ & $\mathbf{AR_{75}}$ & $\mathbf{AR_{M}}$ & $\mathbf{AR_{L}}$ \\ \hline
BU & Ours & 23 & \textbf{69.6} & 86.3 & \textbf{76.6} & \textbf{65.0} & \textbf{76.3} & \textbf{73.5} & 0.881 & \textbf{79.5} & \textbf{68.6} & \textbf{80.3} \\
BU & Newell et al. \cite{Newell2016b} & 6 & 65.5 & \textbf{86.8} & 72.3 & 60.6 & 72.6 & 70.2 & \textbf{89.5} & 76.0 & 64.6 & 78.1 \\
BU & CMU-Pose \cite{Cao2016} & 10 & 61.8 & 84.9 & 67.5 & 57.1 & 68.2 & 66.5 & 87.2 & 71.8 & 60.6 & 74.6 \\ \hline
TD & Megvii \cite{Chen2017a} & - & 73.0 & 91.7 & 80.9 & 69.5 & 78.1 & 79.0 & 95.1 & 85.9 & 74.8 & 84.6 \\
TD & CFN \cite{Huang} & 3 & 72.6 & 86.7 & 69.7 & 78.3 & 64.1 & - & - & - & - & - \\
TD & Mask R-CNN \cite{He2017a} & 5 & 69.2 & 90.4 & 76.0 & 64.9 & 76.3 & 75.2 & 93.7 & 81.1 & 70.3 & 81.8 \\
TD & SJTU \cite{Fang2017} & 0.4 & 68.8 & 87.5 & 75.9 & 64.6 & 75.1 & 73.6 & 91.0 & 79.8 & 68.9 & 80.2 \\
TD & GRMI-2017\footnote{COCO-only results of this entry was obtained from \href{http://presentations.cocodataset.org/Places17-GMRI.pdf}{this talk} on Joint Workshop of the COCO and Places Challenges at ICCV 2017.} \cite{Papandreou2017} & - & 66.9 & 86.4 & 73.6 & 64.0 & 72.0 & 71.6 & 89.2 & 77.6 & 66.1 & 79.1 \\
TD & G-RMI-2016 \cite{Papandreou2017} & - & 60.5 & 82.2 & 66.2 & 57.6 & 66.6 & 66.2 & 86.6 & 71.4 & 61.9 & 72.2 \\ \hline
\end{tabular}}

\end{table}

During ablation experiments we have inspected the effect of different backbones, keypoint detection architectures, and PRN designs. In Table \ref{table:keypoint} and \ref{table:psn} you can see the ablation analysis results on COCO validation set.

\subsubsection{Different Backbones}
We used ResNet models\cite{He2016} as shared backbone to extract features. In Table \ref{table:keypoint}, you can see the impact of deeper features and dilated features. R101 improves the result 1.6 mAP over R50. Dilated convolutions \cite{Chen2016} which is very successful in dense detection tasks increases accuracy 2 mAP over R50 architecture. However, dilated convolutional filters add more computational complexity, consequently hinder realtime performance. We showed that concatenation of $K$ features and intermediate supervision (see Section \ref{sec:kp} for explanations) is crucial for good perfomance. The results demonstrated that performance of our system can be further enhanced with stronger feature extractors like recent ResNext \cite{Xie2016} architectures.

\begin{table}
\begin{center}
\caption{\textbf{Left:} Comparison of different keypoint models. \textbf{Right:} Performance of different backbone architectures. \textit{(no concat: no concatenation, no int: no intermediate supervision, dil: dilated, concat: concatenation)}}
\label{table:keypoint}
\resizebox{0.48\textwidth}{!}{\begin{tabular}{l||ccccc}
\hline
\textbf{Models} & $\mathbf{AP}$ & $\mathbf{AP_{50}}$ & $\mathbf{AP_{75}}$ & $\mathbf{AP_{M}}$ & $\mathbf{AP_{L}}$ \\ \hline
$\text{R}101_{\text{no int.}}$ & 61.3 & 83.7 & 69.6 & 56.6 & 67.4 \\
$\text{R}101_{\text{no concat}}$ & 62.1 & 84.3 & 70.9 & 57.3 & 68.8 \\
$\text{R}101$ & 63.9 & 87.1 & 73.2 & 58.1 & 72.2 \\
$\text{R}101_{\text{dil}}$ & \textbf{64.3} & \textbf{88.2} & \textbf{75} & \textbf{59.6} & \textbf{73.9} \\ \hline
\end{tabular}}
\quad
\resizebox{0.48\textwidth}{!}{\begin{tabular}{l||ccccc}
\hline
\textbf{Backbones} & $\mathbf{AP}$ & $\mathbf{AP_{50}}$ & $\mathbf{AP_{75}}$ & $\mathbf{AP_{M}}$ & $\mathbf{AP_{L}}$ \\ \hline
R50 & 62.3 & 86.2 & 71.9 & 57.7 & 70.4 \\
R101 & 63.9 & 87.1 & 73.2 & 58.1 & 72.2 \\
$\text{R}101_{\text{dil}}$ & \textbf{64.3} & \textbf{88.2} & \textbf{75} & \textbf{59.6} & \textbf{73.9} \\ \hline
\end{tabular}}
\end{center}
\end{table}

\subsubsection{Different Keypoint Architectures}
Keypoint estimation requires dense prediction over spatial locations, so its performance is dependent on input and output resolution. In our experiments, we used $480 \times 480$ images as inputs and outputted $120 \times 120 \times (K+1)$ heatmaps per input. $K$ is equal to 17 for COCO dataset. The lower resolutions harmed the mAP results while higher resolutions yielded longer training and inference complexity. We have listed the results of different keypoint models in Table \ref{table:keypoint}.

The intermediate loss which is appended to the outputs of $K$ block’s enhanced the precision significantly. Intermediate supervision acts as a refinement process among the hierarchies of features. As previously shown in \cite{Cao2016,Newella,Wei2016}, it is an essential strategy in most of the dense detection tasks.

We have applied a final loss to the concatenated $D$ features which is downsized from K features. This additional stage ensured us to combine multi-level features and compress them into a uniform space while extracting more semantic features. This strategy brought 2 mAP gain in our experiments. 
\subsubsection{Pose Residual Network Design} 
PRN is a simple yet effective assignment strategy, and is designed for faster inference while giving reasonable accuracy. To design an accurate model we have tried different configurations. Different PRN models and corresponding results can be seen in Table \ref{table:psn}. These results indicate the scores obtained from the assignment of ground truth person bounding boxes and keypoints.

\begin{table}
\begin{center}
\caption{\textbf{Left:} Performance of different PRN models on COCO validation set. \textit{N: nodes, D: dropout and R: residual connection.}  \textbf{Right:} Ablation experiments of PRN with COCO validation data.}
\label{table:psn}
\resizebox{0.48\textwidth}{!}{\begin{tabular}{l||ccccc}
\hline
\textbf{PRN Models} & $\mathbf{AP}$ & $\mathbf{AP_{50}}$ & $\mathbf{AP_{75}}$ & $\mathbf{AP_{M}}$ & $\mathbf{AP_{L}}$ \\ \hline
1 Layer 50 N & 76.3 & 89.2 & 79.1 & 74.8 & 80.4 \\
1 Layer 50 N, D & 78.6 & 91.7 & 82.4 & 77.1 & 83.1 \\
1 Layer 512 N, D & 84.1 & 94.2 & 85.3 & 82 & 86.2 \\
2 Layers 512 N, D & 81.9 & 91.1 & 82.6 & 79.8 & 84.3 \\
1 Layer 2048 N, D+R & \text{83.2} & \text{95.7} & \text{86.1} & \text{82.0} & \text{86.3} \\
1 Layer 1024 N, D+R & \textbf{89.4} & \textbf{97.1} & \textbf{91.2} & \textbf{87.9} & \textbf{91.8} \\ \hline
\end{tabular}}
\quad
\resizebox{0.48\textwidth}{!}{\begin{tabular}{l||ccccc}
\hline
\textbf{PRN Ablations} & $\mathbf{AP}$ & $\mathbf{AP_{50}}$ & $\mathbf{AP_{75}}$ & $\mathbf{AP_{M}}$ & $\mathbf{AP_{L}}$ \\ \hline
Both GT & 89.4 & 97.1 & 91.2 & 87.9 & 91.8 \\
GT keypoints + Our bbox & 75.3 & 82.1 & 78 & 70.1 & 84.5 \\
Our keypoints + GT bbox & 65.1 & 89.2 & 76.2 & 60.3 & 74.7 \\
PRN & 64.3 & 88.2 & 75 & 59.6 & 73.9 \\
UCR & 49.7 & 59.5 & 52.4 & 44.1 & 51.6 \\
Max & 45.3 & 55.1 & 48.8 & 40.6 & 46.9 \\ \hline
\end{tabular}}
\end{center}
\end{table}

We started  with a primitive model which is a single hidden-layer MLP with 50 nodes, and added more nodes, regularization and different connection types to balance speed and accuracy. We found that 1024 nodes MLP, dropout with 0.5 probability and residual connection between input and output boosts the PRN performance up to $89.4$ mAP on ground truth inputs.


\begin{table}
\begin{center}
\caption{PRN assignment results with non-grouped keypoints obtained from two bottom-up methods.}
\label{table:prnother}
\resizebox{0.48\textwidth}{!}{\begin{tabular}{l||ccccc}
\hline
\textbf{Models} & $\mathbf{AP}$ & $\mathbf{AP_{50}}$ & $\mathbf{AP_{75}}$ & $\mathbf{AP_{M}}$ & $\mathbf{AP_{L}}$ \\ \hline
Cao et al. \cite{Cao2016} & 58.4 & 81.5 & 62.6 & \textbf{54.4} & 65.1 \\
PRN + \cite{Cao2016} & \textbf{59.2} & \textbf{82.2} & \textbf{64.4} & 54.1 & \textbf{67.0} \\ \hline
Newell et al. \cite{Newell2016b} & 56.9 & 80.8 & 61.3 & 49.9 & \textbf{68.8} \\
PRN + \cite{Newell2016b} & \textbf{58.1} & \textbf{81.4} & \textbf{63.0} & \textbf{51.3} & 68.1 \\ \hline
\end{tabular}}
\end{center}
\end{table}

In ablation analysis of PRN (see Table \ref{table:psn}), we compared \textit{Max}, \textit{UCR} and \textit{PRN} implementations (see Section \ref{sec:method-prn} for descriptions) along with the performance of PRN with ground truth detections. We found that , lower order grouping methods could not  handle overlapping detections, both of them performed poorly. As we hypothesized, PRN could overcome ambiguities by learning meaningful pose structures (Fig. \ref{fig:psnatom} (right)) and improved the results by $\sim$20 mAP over naive assignment techniques. We evaluated the impact of keypoint and person subnets to the final results by alternating inputs of PRN with ground truth detections. With ground truth keypoints and our person detections, we got 75.3 mAP,  it shows that there is a large room for improvement in the keypoint localization part. With our keypoints and ground truth person detections, we got 65.1 mAP. This can be interpreted as our person detection subnet is performing quite well. Both ground truth detections got 89.4 mAP, which is a good indicator of PRN performance. In addition to these experiments, we tested PRN on the keypoints detected by previous SOTA bottom-up models \cite{Cao2016,Newell2016b}. Consequently, PRN performed better grouping (see Table \ref{table:prnother}) than their methods: \textit{Part Affinity Fields}\cite{Cao2016} and \textit{Associative Embedding}\cite{Newell2016b} by improving both detection results by $\sim$1 mAP. To obtain results in Table \ref{table:prnother}, we have used COCO val split, our person bounding box results and the keypoint results from the official source code of the papers.     Note that running PRN on keypoints that were not generated by MultiPoseNet is unfair to PRN because it is trained with our detection architecture. Moreover original methods use image features for assignment coupled with their detection scheme, nonetheless, PRN is able to outperform the other grouping methods.    
\subsection{Person Detection}
We trained the person detection subnet only on COCO person instances by freezing the backbone with keypoint detection parameters. The person category results of our network with different backbones can be seen in Table \ref{table:person}. We compared our results with the original methods that we adopt in our architecture. Our model with both ResNet-50 and ResNet-101 backends outperformed the original implementations. This is not a surprising result since our network is only dealing with a single class whereas the original implementations handle $80$ object classes.

\begin{table}[h]
\begin{center}
\caption{\textbf{Left:} Person detection results on COCO dataset. \textbf{Right:}Person segmentation results on PASCAL VOC 2012 test split.}
\label{table:person}
\begin{tabular}{l||cccccc}
\hline
\textbf{Person Detectors} & $\mathbf{AP}$ & $\mathbf{AP_{50}}$ & $\mathbf{AP_{75}}$ & $\mathbf{AP_{S}}$ & $\mathbf{AP_{M}}$ & $\mathbf{AP_{L}}$ \\ \hline
Ours - R101 & \textbf{52.5} & \textbf{81.5} & \textbf{55.3} & \textbf{35.2} & \textbf{59} & 71 \\
Ours - R50 & 51.3 & 81.4 & 53.6 & 34.9 & 58 & 68.1 \\
RetinaNet \cite{Lin2017} & 50.2 & 77.7 & 53.5 & 31.6 & \textbf{59} & \textbf{71.5} \\
FPN \cite{Lina} & 47.5 & 78 & 50.7 & 28.6 & 55 & 67.4 \\ \hline
\end{tabular}
\quad
\begin{tabular}{l||c}
\hline
\textbf{Segmentation} & \textbf{IoU} \\ \hline
DeepLab v3 \cite{Chen} & \textbf{92.1} \\ 
DeepLab v2 \cite{Chen2016} & 87.4 \\ 
SegNet \cite{kendall2015bayesian} & 74.9 \\ \hline
Ours & 87.8 \\ \hline
\end{tabular}
\end{center}
\end{table}
\subsection{Person Segmentation}
\begin{figure}
\centering
\includegraphics[width=\textwidth]{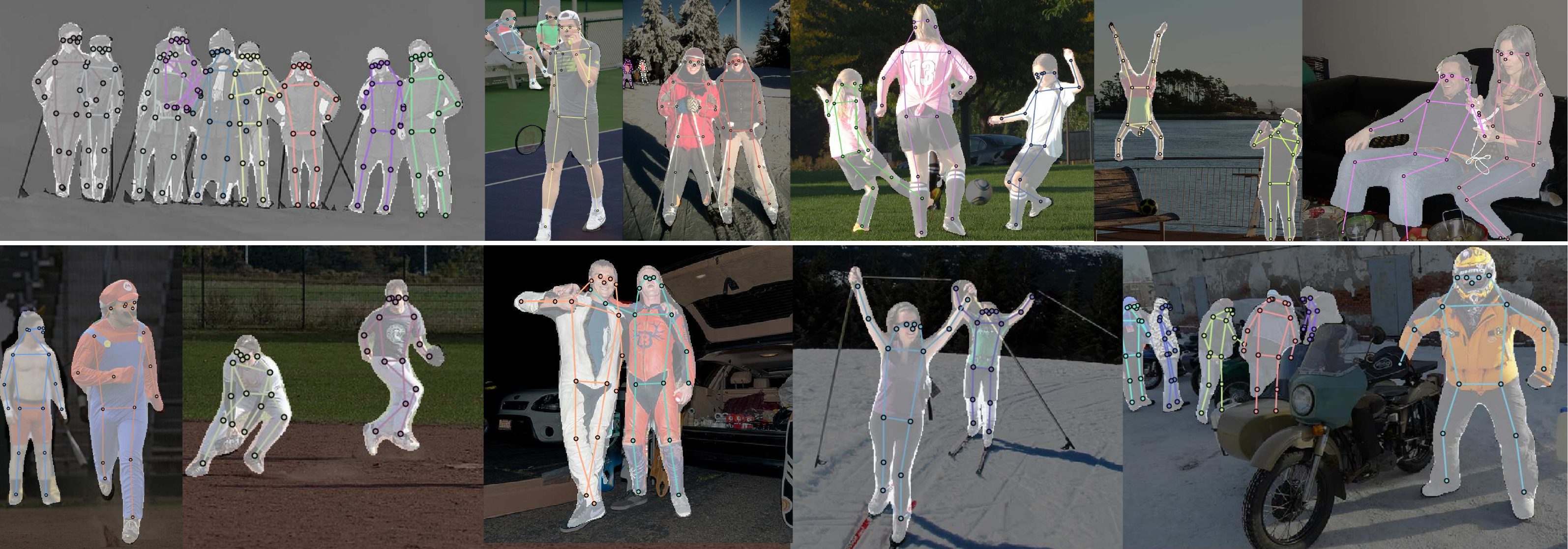}
\caption{Some qualitative results for COCO test-dev dataset.}
\label{fig:samples}
\end{figure}

Person segmentation output is an additional layer appended to the keypoint outputs. We obtained the ground truth labels by combining person masks into single binary mask layer, and we jointly trained segmentation with keypoint task. Therefore, it adds a very small complexity to the model. Evaluation was performed on PASCAL VOC 2012 test set with PASCAL IoU metric. We obtained final segmentation results via multi-scale testing and thresholding. We did not apply any additional test-time augmentation or ensembling. Table \ref{table:person} shows the test results of our system in comparison with previous successful semantic segmentation algorithms. Our model outperformed most of the successful baseline models such as SegNet \cite{kendall2015bayesian} and Deeplab-v2 \cite{Chen2016}, and got comparable performance to the state-of-the-art Deeplab v3 \cite{Chen} model. This demonstrates the capacity of our model to handle different tasks altogether with competitive performance. Some qualitative segmentation results are given in Fig. \ref{fig:samples}.

\subsection{Runtime Analysis}
\begin{figure}[!tbp]
  \centering
  \begin{minipage}[b]{0.47\textwidth}
    \includegraphics[width=\textwidth]{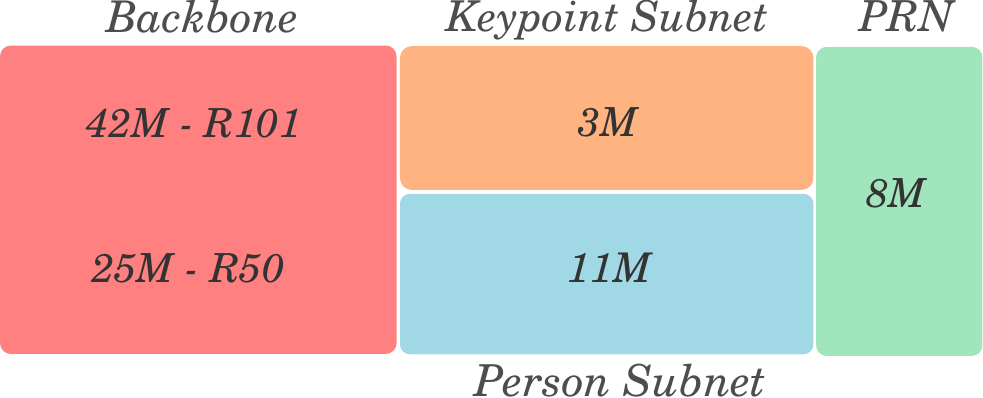}
    \caption{Number of parameters for each block of MultiPoseNet.}
   \label{fig:params}
  \end{minipage}
  \hfill
  \begin{minipage}[b]{0.47\textwidth}
    \includegraphics[width=\textwidth]{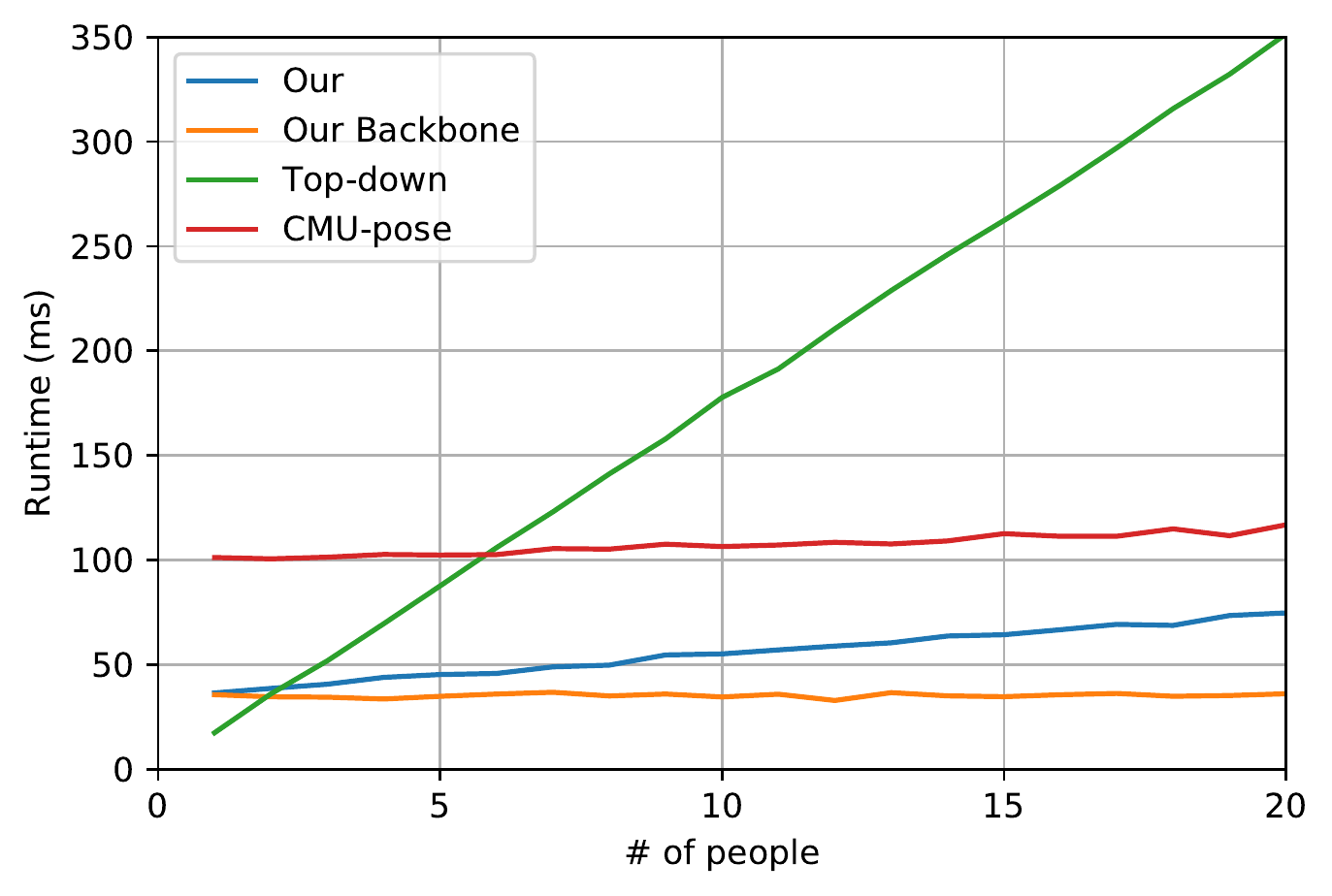}
    \caption{Runtime analysis of MultiPoseNet with respect to number of people.}
   \label{fig:runtime}
  \end{minipage}
\end{figure}

Our system consists of a backbone, keypoint \& person detection subnets, and the pose residual network. The parameter sizes of each block is given in Fig. \ref{fig:params}. Most of the parameters are required to extract features in the backbone network, subnets and PRN are relatively lightweight networks. So most of the computation time is spent on the feature extraction stage. By using a shallow feature extractor like ResNet-50, we can achieve realtime performance. To measure the performance, we have built a model using ResNet-50 with $384 \times 576$ sized inputs which contain $1$ to $20$ people. We measured the time spent during the inference of 1000 images, and averaged the inference times to get a consistent result (see Fig. \ref{fig:runtime}). Keypoint and person detections take 35 ms while PRN takes 2 ms per instance. So, our model can perform between 27 (1 person) and 15 (20 persons) FPS depending on the number of people.

\section{Conclusion} 
In this work, we introduced the Pose Residual Network that is able to accurately assign keypoints to  person detections outputted by a multi task learning architecture (MultiPoseNet). Our pose estimation method achieved state-of-the-art performance among bottom-up methods and comparable results with top-down methods. Our method has  the fastest inference time compared to previous methods.  We showed the assignment performance of pose residual network  ablation analysis. We demonstrated the representational capacity of our multi-task learning model by jointly producing keypoints, person bounding boxes and person segmentation results.
\pagebreak
\bibliographystyle{splncs}
\bibliography{references}
\end{document}